\documentclass[letterpaper, 10 pt, conference]{ieeeconf}  % Comment this line out if you need a4paper

\IEEEoverridecommandlockouts                              % This command is only needed if 
\usepackage{times}
\usepackage{epsfig}
\usepackage{graphicx}
\usepackage{amsmath}
\usepackage{amssymb}
\usepackage{xcolor}
\usepackage{tabularx}
\usepackage{rotating}
\usepackage{verbatim}
\usepackage{xcolor,colortbl}
\usepackage{stackengine}
\usepackage{gensymb}
\usepackage{booktabs}
\usepackage{subcaption}
\usepackage{algorithm}
\usepackage[noend]{algpseudocode}
\usepackage{multicol}

\algrenewcommand\algorithmicforall{\textbf{foreach}}
\algrenewcommand\algorithmicindent{.8em}

\title{\LARGE \bf
Goal-constrained Sparse Reinforcement Learning \\
for End-to-End Driving
}

\author{Pranav Agarwal$^{1}$, Pierre de Beaucorps$^{1}$ and Raoul de Charette$^{1}$% <-this % stops a space
\thanks{$^{1}$Pranav Agarwal, Pierre de Beaucorps and Raoul de Charette are from Inria, Paris.
        {\tt\small first.last-name@inria.fr}}%
}

\begin{document}

\maketitle
\thispagestyle{empty}
\pagestyle{empty}
	
%%%%%%%%% ABSTRACT
\begin{abstract}
Deep reinforcement Learning for end-to-end driving is limited by the need of complex reward engineering. Sparse rewards can circumvent this challenge but suffers from long training time and leads to sub-optimal policy. In this work, we explore full-control driving with only goal-constrained sparse reward and propose a curriculum learning approach for end-to-end driving using only navigation view maps that benefit from small virtual-to-real domain gap. To address the complexity of multiple driving policies, we learn concurrent individual policies selected at inference by a navigation system. We demonstrate the ability of our proposal to generalize on unseen road layout, and to drive significantly longer than in the training.
\end{abstract}

%%%%%%%%% BODY TEXT
%!TEX root = main.tex

\section{Introduction}
\label{sec:intro}

Deep reinforcement learning (RL) has successfully solved some of the most challenging tasks and off the charts performance in games like Atari~\cite{Mnih2013PlayingAW}, Go~\cite{Silver2016MasteringTG} and Dota 2~\cite{Berner2019Dota2W} has shown its potential to solve complex decision making problems even for long term gains. Considering these results, RL has been used to solve real life robotics applications like manipulation~\cite{Gu2017DeepRL,Kalashnikov2018QTOptSD} and autonomous driving~\cite{You2017VirtualTR,Chen2019ModelfreeDR}. However, most of the approaches are domain dependent and cannot be generalized across tasks.

Traditionally rule-based approach addressing tasks like autonomous driving requires complex engineering~\cite{Katrakazas2015RealtimeMP} and greedy search based algorithms~\cite{Mohanty2013CuckooSA}.
Learning to drive with reinforcement learning (RL) is an alternative since the optimal policy is optimized from reward.
However, RL commonly employs reward shaping~\cite{Ng1999PolicyIU, Laud2003TheIO, TenorioGonzlez2010DynamicRS, Randlv1998LearningTD} -- an extensive manual tuning subsequently prone to human bias~\cite{Hu2020LearningTU}. In real life, humans are often rewarded \textit{only} when the task is complete. 
Sparse reward follows the same analogy and is domain independent, hence easily transferred to new tasks. However, lack of feedback signals makes sparse reward RL difficult to train~\cite{Hare2019DealingWS,wang2020deep}, and to the best of our knowledge no work yet address end-to-end driving with sparse reward. Only~\cite{rafati2019learning,hare2019dealing} experiment longitudinal driving task in simulation, without any lateral action.

\begin{figure}
    \centering
    \includegraphics[width=1.0\linewidth]{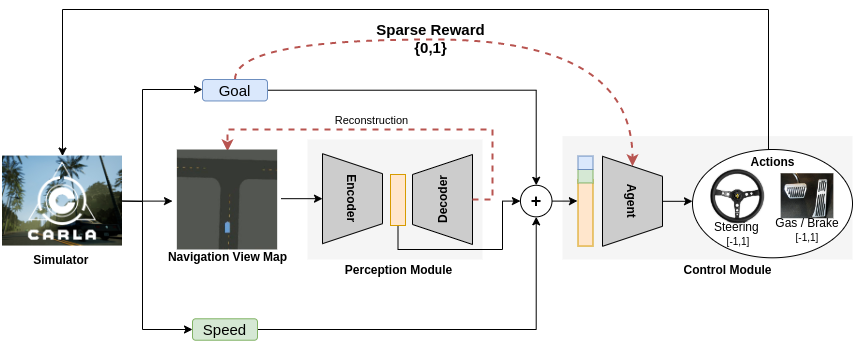}
    \caption{\textbf{End to End Driving.} Overview of our reinforcement learning framework for end to end driving with sparse reward. The perception module encodes the navigation view map and the control module learns the optimal low level control (throttle, steering, brake) from the world model, current speed, distance covered and the goal.}
    \label{fig:overview}
\end{figure}

We instead address this lack and learn the full car control with a goal-constrained sparse reinforcement learning framework, shown in Fig.~\ref{fig:overview}.
Because driving forms a natural curriculum where difficulty scales with the traveled distance, we train our method in a curriculum learning fashion with goals of increasing complexity. 
However, sparse reward can hardly train a single policy to maneuver all road scenarios (left turn, driving straight, round-about, etc.) which require different skills. 
We tackle the above problem by designing a multi-policies strategy where individual policies are learned concurrently. 
At inference, a navigation system handles the decision at intersections (e.g. which direction to go), subsequently switching to the ad-hoc policy.
Instead of the often used front-view images \cite{ohn2020learning,Jaritz2018EndtoEndRD,Kiran2020DeepRL} or multi-modal observation space~\cite{prakash2021multi}, we train using \textit{only} navigation view maps and learn a compact world model~\cite{Ha2018WorldM} from the latter. Both world model (VAE) and RL policy (Proximal Policy Algorithm~\cite{Schulman2017ProximalPO}) are trained simultaneously. 
Considering that with sparse reward the RL policy is trained online where the latent space used is optimized after each episode, the task is very challenging. Hence, we experiment in a simulated environment without other road users, already significantly more complex than previous attempts~\cite{rafati2019learning,hare2019dealing}. The main contribution of our work are: 
\vspace{-0.05em}
\begin{itemize}
    \item \textbf{{RL driving with navigation maps only.}} Instead of front-view images, we leverage navigation maps, having smaller domain gaps and better generalization capacity, 
    \item \textbf{{Policies from goal-constrained binary sparse reward.}} 
    Benefiting from curriculum learning and our revert strategy, we learn the \textit{full} car control of individual driving policies with goal-constrained binary sparse reward,
    \item \textbf{{Flexible perception and control learning.}} 
    Our perception and control modules can be trained sequentially or \textit{simultaneously}, therefore balancing the benefit of policy training from pixels or from a compact world model.
\end{itemize}
%!TEX root = main.tex
% \pagebreak
\section{Related Works}
\label{sec:related}

\noindent\textbf{End-to-end driving with RL.}
As it requires a trial-and-error process, the common RL strategy involves training virtual agents with simple rewards, in a model-based RL fashion.
Reward shaping is of the utmost importance since it provides a supervision signal for the network, and was subsequently extensively discussed in~\cite{Nageshrao2019AutonomousHD,Chen2020InterpretableEU,Lillicrap2016ContinuousCW,Kendall2019LearningTD}.
Dense (i.e. frame-wise) reward are commonly applied, by penalizing velocity~\cite{Kendall2019LearningTD,Li2019UrbanDW} along with road-car angle~\cite{mnih2016asynchronous}, distance to center~\cite{Jaritz2018EndtoEndRD,Lillicrap2016ContinuousCW,toromanoff2019deep,Meghjani2019ContextAI}, distance from destination \cite{Dosovitskiy2017CARLAAO}, car collisions~\cite{Dosovitskiy2017CARLAAO}. Some works also optimize comfort accounting for steering and throttle \cite{Chen2019ModelfreeDR} and traffic rules \cite{Li2019UrbanDW}. 
The main limitation of dense reward is that it acts as an expert to imitate, limiting the discovery~\cite{Marom2018BeliefRS} and preventing human-like behavior -- like taking a turn from inside~\cite{Jaritz2018EndtoEndRD}. We refer to the recent survey \cite{tampuu2020survey} for details.\\
On input representations, most RL methods learn directly control mapping from front-view cameras~\cite{Lillicrap2016ContinuousCW,toromanoff2019deep,Jaritz2018EndtoEndRD,toromanoff2018end} though 
\cite{pan2017virtual} relates that this limits the transferability to real world. Rare works employ navigation maps and only for imitation learning~\cite{hecker2020learning}, or robot navigation~\cite{chen2020robot}.\\
Different from the literature, we learn multiple individual policies for the end-to-end driving task from navigation maps only, thus benefiting from higher generalization capability.

\noindent{}\textbf{Sparse reward RL.} Relying on sparse reward is difficult to optimize, hence often applied in conjunction of other training strategies. For example, \cite{trott2019keeping,riedmiller2018learning} use auxiliary tasks to prevent suboptimal policy and encourage exploration. The latter can be employed as a meta reward learning strategy \cite{agarwal2019learning}. 
Intrinsic rewards are also used, in the form of sub-goals \cite{dann2019deriving} and demonstrations \cite{vecerik2017leveraging}. Of interest, note that to ease the binary reward feedback, exploration is enforced which is mostly suitable in finite observation space.
Instead, driving has an infinite observation space meaning to unwanted exploration further slowing the training.
To leverage some exploration despite observation space, we employ goal-constrained curriculum learning which we now review.

\noindent{}\textbf{Curriculum learning.}
Curriculum is traditionally employed~\cite{Elman1993LearningAD, Rohde1999LanguageAI, Sanger1994NeuralNL} as a way to break down complex tasks into simple ones of increasing complexities~\cite{Lazaric2008TransferOS, Taylor2009TransferLF} -- similarly to human learning~\cite{Khan2011HowDH}. It has been reintroduced for deep learning~\cite{Bengio2009CurriculumL}, leading to many extensions~\cite{Weinshall2018CurriculumLB,Matiisen2020TeacherStudentCL,Graves2017AutomatedCL,Portelas2019TeacherAF}. 
Controlling the complexity of curriculum is the key element and sophisticated strategies are used like teacher guided~\cite{Graves2017AutomatedCL} , self play \cite{Sukhbaatar2018IntrinsicMA} or the use of GANs for goal generation~\cite{Held2018AutomaticGG, Racanire2019AutomatedCT}.

To the best of our knowledge, no work yet addressed driving from sparse reward, possibly relating to the complexity of sparse reward in long-horizon tasks~\cite{jiang2019hierarchical} and structured environment like road layout. 
However, sparse reward main benefit is to let the network discover efficient driving strategy. 
We now present our method addressing end-to-end driving with RL from sparse binary reward in a curriculum fashion.
%!TEX root = main.tex

% \begin{figure}
%     \centering
%     \includegraphics[width=0.6\linewidth]{figure/icra_5.png}
%     \caption{\textbf{Overview of the network architecture of perception and control modules.} The arrows show the flow of gradients during training. The encodings learnt by the encoder are input to actor and critic which is concatenated with the speed and curriculum for a given time step and episode respectively.}
%     \label{fig:architecture}
% \end{figure}

% \begin{figure*}
%     \centering
%     \includegraphics[width=1.0\linewidth]{figure/icra_15.png}
%     \caption{\textbf{Reconstruction of the VAE perception module.} The reconstructed images (right) using the latent variable and the original image (left). With several training iterations the  embeddings are optimised in order to optimally compress the original image.}
%     \label{fig:architecture}
% \end{figure*}

% \begin{figure}
%     \centering
%     \includegraphics[width=0.65\linewidth]{figure/icra_6.png}
%     \caption{\textbf{Constraint dependent reward strategy.} The agent receives a reward if it reaches the current goal, that is covers C meters, in the given time and at the goal waypoint it satisfies the distance as well as the orientation constraint, that is, the agent's location is within the green patch ($\Delta$x meters from the centre) and its orientation with respect to the next waypoint does not exceed the $\Delta \theta$ range.}
%     \label{fig:architecture}
% \end{figure}

\section{Method}
\label{sec:method}
Fig.~\ref{fig:overview} shows an overview of our reinforcement learning (RL) framework, where the agent interacts with Carla~\cite{Dosovitskiy2017CARLAAO} virtual driving environment and predicts the continuous low level control of a car (steering, throttle, brake).
There are important originalities of our proposal regarding the literature.
Firstly, rather than front view images we rely on navigation views (i.e. top-view navigation maps) which prove smaller virtual-to-real gap. 
Secondly, to avoid the pitfall of model-based RL --~where agents mimic the model, we only provide a sparse binary reward at the end of each episode ; 1 if the goal is reached, 0 if not. This light supervision enables the discovery of unique driving features.
Thirdly, we learn a compact world model enabling better policy learning, and train the whole pipeline in a curriculum fashion for separate policies (driving straight, turning left, turning right).
For inference a simple navigation system apply policy switching.

% Learning a complex task like driving using sparse reward is challenging for a reinforcement learning agent. Large observation space further complicates the process with the increase in the number of trainable parameters. 
The remaining of this section describes the architecture (Sec.~\ref{sec:meth-architecture}), our sparse reinforcement learning pipeline (Sec.~\ref{sec:meth-PPO}), and the training strategy (Sec.~\ref{sec:meth-multipolicies}).
% end to end framework for learning to drive using sparse rewards along with a hybrid learning approach to learn an efficient policy using high dimensional inputs.
%\RC{Many details are missing in the method description. You're skipping many general insights: I wouldn't understand from the current    text. Go from general to details}

\subsection{Architecture}\label{sec:meth-architecture}
%\RC{What network, describe the VAE, how you concat the encoding, etc. Use figure highlights in the text.}
Our architecture is composed of two main modules: \textit{perception} which provides a compact representation of the world, and \textit{control} which predicts the full car control output.

The perception module takes as input a single navigation view map (256x256 RGB) output by Carla simulator. It is composed of a shallow Variational AutoEncoder (VAE) having 5 convs with increasing filters (32,64,128,256,256) as encoder, and 5 transpose convs (256,128,64,32,3) in the decoder. The latent space is a 256 dimension vector, hereafter referred as \textit{world model}. 
% (5 layers with 32, 64, 128, 256, 256 filters with a common kernel size of 4 and strides of 2 with same padding and a relu non-linearity)  (encoder) and transpose convolutional (5 layers with 256, 128, 64, 32, 3 filters, other parameters similar to encoder) (deocder) layers. 

The control module takes as input the concatenation of the world model, with current speed and goal easily retrieved from Carla metadata, as illustrated in Fig.~\ref{fig:overview}. 
It employs a shallow network consisting of 4 FC layers (512,256,128,64 neurons) and a 2 neuron output layer with tanh activation to ouput steering and brake/throttle as scalars in~$[-1, +1]$. The prediction is logically applied to the simulator before obtaining the next simulation step.
% For every frame this world model outputs a 256 dimensional encodings and are further concatenate with the speed, current location and the goal location of the agent. 
% These are then used to output a low level control following the PPO algorithm.

\subsection{Sparse reinforcement learning}\label{sec:meth-sparsecur}

% To learn ad-hoc control, the literature employs dense reward (i.e. frame-wise) penalizing the agent as it differs from a model -- typically, driving in the center of the lane -- subsequently constraining the discovery of driving styles (e.g. taking a turn from the inside). 
Instead of penalizing the agent with dense frame-wise reward, we define a goal to reach and only reward the agent (+1) if goal is reached before the episode termination. 
As such, our supervision signal is significantly weaker than existing dense reward end-to-end driving~\cite{tampuu2020survey}, but encourages self discovery. 
As other long-horizon tasks~\cite{jiang2019hierarchical}, learning driving is indubitably hard with a sparse reward. We employ thus curriculum learning strategy to break complexity, and use an on-policy Proximal Policy Optimization (PPO) as it is less sensitive to hyperparameters tuning while ensuring slow policy deviation during training. Given the changing goal distribution with curriculum, on-policy algorithm is more efficient than an off-policy, and recent work \cite{yu2021surprising} demonstrated the benefit of PPOs for complex non-stationary environment.
%Moreover a recent work \cite{yu2021surprising} has successfully shown PPOs to be more effective in solving complex non stationary environments.} 
We now provide the reader with a brief background.
\\

% Learning from pixels has several advantage but it is hard to optimise. So we learn a world model as introduced in \cite{Ha2018WorldM} for the different navigation maps of a given Carla environment. Learning a prior world model to train a policy limits the encodings to the current route only, while we introduce a lifelong learning approach where the embeddings are continuously optimised based on the given route. This breaks the need to retrain the world every time the policy is being trained on a new route or if multiple policies being trained on different routes. With this approach we have an end to end framework which has an advanatge of learning directly from pixels and learning an optimal policy using low dimensional encodings. 

\noindent\textbf{Proximal Policy Optimization (PPO).}\label{sec:meth-PPO} 
% We just provide the reader with a background understanding of PPO on-policy algorithm~ \cite{Schulman2017ProximalPO}. 
% We chose PPO~\cite{Schulman2017ProximalPO} for its stable performance. 
In short, considering a state $s_{t}$ and an action space $a_{t}$, PPO~\cite{Schulman2017ProximalPO} optimizes network parameters $\theta$ using a surrogate objective $L(.)$ with a stochastic gradient ascent to learn the best policy $\pi$, while avoiding large deviation from last policy $\pi _{\theta _{old}}$. This writes
% as shown in equation \eqref{eq:1} which prevents performance collapse as observed in previous policy gradient algorithms like REINFORCE \cite{Williams2004SimpleSG} and more sample efficient. 
\begin{equation}
L(\theta ) = E_{t}\left [ \frac{\pi _{\theta}\left ( a_{t}|s_{t} \right )}{\pi _{\theta_{old}}\left ( a_{t}|s_{t} \right )} A_{t}^{\pi _{\theta _{old}}}\right ]\,,
\label{eq:1}
\end{equation}
% Here, $\pi$ is the policy used to map the observation space ($s_{t}$) to its action space ($a_{t}$).
with $A_{t}$ being the advantage of the predicted action -- that is, \textit{How good/bad was the predicted action}.

% PPO also has an added advantage of being stable and easy to implement. Taking all these factors into account, we use it for learning from sparse rewards.
While it was originally proposed with either KL penalty or clipped objective to solve the constrained policy optimization problem, we use the lighter clipped version which writes 
\begin{equation}
    L^{\text{clip}}\left ( \theta  \right ) = E_{t}\left [ \min\left ( r_{t}\left ( \theta  \right )A_{t},\text{clip}\left ( r_{t}\left ( \theta  \right ),1 - \epsilon , 1 + \epsilon  \right )A_{t}\right ) \right ]\label{eq:2}\,,
\end{equation}
with reward $r_{t}$ being function of current and old policy,
\begin{equation}
r_{t} = \frac{\pi_{\theta}\left ( a_{t}|s_{t} \right )}{\pi_{\theta_{old}}\left ( a_{t}|s_{t} \right )}\,,
\end{equation}
bounded by $\epsilon$ within a range of $[1-\epsilon,1+\epsilon]$. This prevents the new policy $\pi_{\theta}$ from large changes as compared to the old policy $\pi_{\theta_{old}}$.
To further reduce variance, the generalized advantage estimation $A_{t}$ is the weighted average of several advantage functions.\\

% \RC{1. Sparse reward reinforcement learning}
% \RC{2. Curriculum learning}
% \RC{3. Hybrid VAE/RL training (algorithm, hybrid VAE/RL training, etc.)}
% \subsection{Hybrid Perception/Policy}

% In this section, we describe our end to end learning framework to learn to drive using sparse rewards. Our framework consists of two separate modules the perception module and the control module, each being trained simultaneously. The perception module is an encoder decoder architecture consisting of stacks of convolution and transpose convolution layers. The main aim is to compress the high dimensional image of the world, which further stabilises policy training. This network is trained using data from a buffer collected at every simulation step. Previous work learnt the world model prior to training the policy. This limited the range of embeddings to a particular dataset only. Any modification in the input data would require training from scratch. Our approach overcomes this disadvantage by learning both the modules simultaneously. 

\begin{figure}
    \centering
    \subfloat[Curriculum revert strategy]{\includegraphics[width=0.6\linewidth, height=0.25\linewidth]{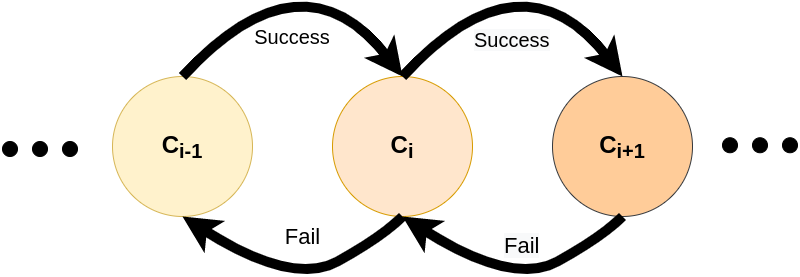}\label{fig:Curriculum}}\hfill
    \subfloat[Goal constraints]{\includegraphics[width=0.34\linewidth,height=0.34\linewidth]{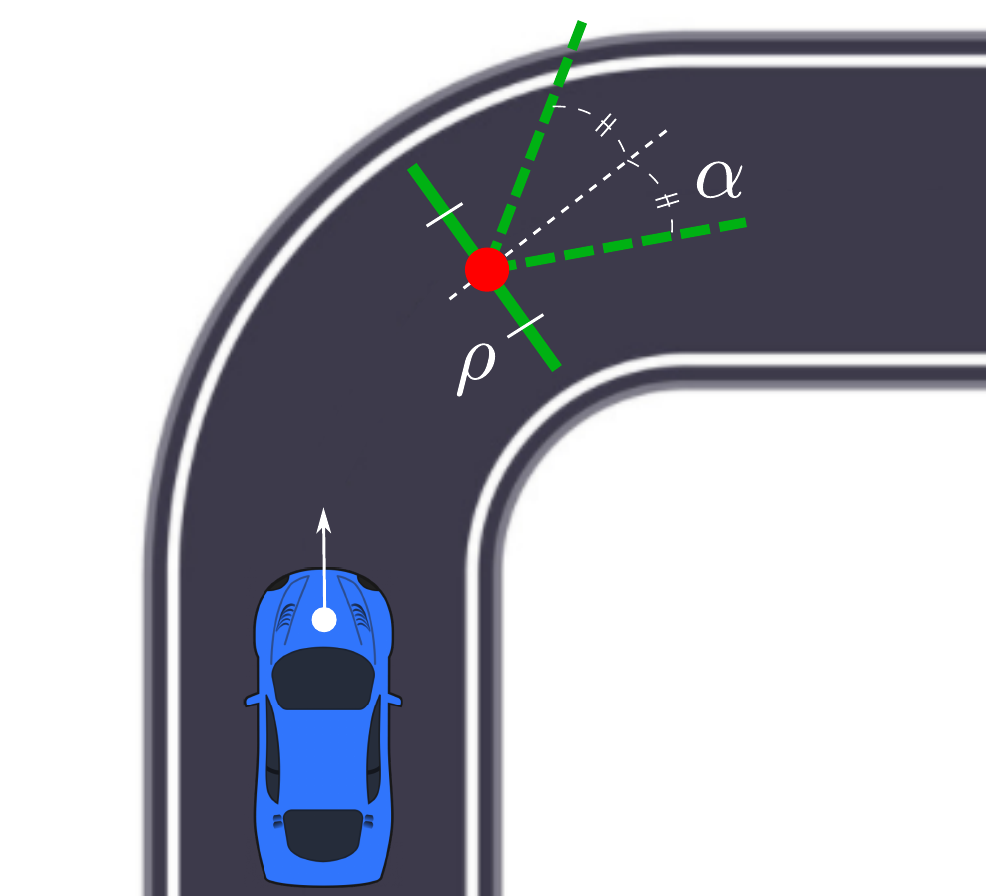}\label{fig:goal}}
    \caption{\textbf{Curriculum strategy and Goal constraints.} (a)~Instead of simply increasing complexity, we introduce a revert mechanism upon failure. (b) is a schematic view of a goal (red) and our constraints (green). Details in text.}
\end{figure}

\noindent\textbf{Curriculum learning.}
Unlike dense ones, sparse rewards are known to be \textit{sample inefficient} because of the little guidance provided by the reward.
To fight this, we use curriculum learning~\cite{Bengio2009CurriculumL} where \textit{complexity} is the distance the agent needs to drive. 
We customize the curriculum setup for our problem.

Firstly, instead of monotonically increasing the complexity, we introduce a \textit{revert strategy} -- inspired by~\cite{Dasagi2019CtrlZRF}. While the latter saved checkpoint weights and restored them on-demand, our strategy is to increase complexity only when the agent succeed and decrease it upon failure, as illustrated in Fig.~\ref{fig:Curriculum}. We argue that similarly to~\cite{Dasagi2019CtrlZRF}, this prevents policy to remain stuck in a local minimum. 
Empirically, we observe that it plays an important role in the training (see Sec.~\ref{sec:experiments}).
For each episode, the goal is defined as a 2 DoF (Degree of Freedom) vector being the target position.
% We start with a simple goal of 1 meter and for every success goal is incremented by 1 meter and in case of failure it is reduced by 1 meter as shown in fig.~\ref{fig:Curriculum}. 
% For simple policies like straight driving the rule for increasing the complexity can be varied, while for more complex task like left and right turn we observe best performance with 1 meter increment.
%\RC{More details}
% Though we can train it for long routes we find that training the agent for a 100 meter complexity is enough to optimize to the best policy.

% This approach of learning to drive using sparse rewards provides an external agent to integrate complex driving maneuvers  and optimise the original driving in the form of constraints to the original reward strategy. 
Secondly, we apply goal constraints by introducing a lateral radius~$\rho$ to the goal position, and a maximum angle difference~$\alpha$ w.r.t. the local road curvature.
Specifically, the goal is being reached when the agent position is within $[-\rho, +\rho]$ of the goal position \textit{and} if its relative road orientation is $[-\alpha,+\alpha]$. 
The intuition is that it forces the agent to cross a virtual 'finish line' with a certain orientation as illustrated in Fig.~\ref{fig:goal}, logically boosting policy for the next complexity. We set $\rho=1m$ and $\alpha=15\degree$.

In practice, we start the curriculum with a complexity of 1 (i.e. 1m goal) and increment/decrement it by 1 according to our curriculum. 
Upon goal completion the agent is rewarded `+1' and the episode ends.
If the goal is not reached before the episode terminates -- without any reward. Importantly, we stop the training after 100m goal distance is reached.
% We emphasize here that sparse learning allows future work to account for traffic rules, yielding, overtaking, etc.

\subsection{Training strategy}\label{sec:meth-multipolicies}
% In order to learn an optimal policy or perception encoding covering different aspects of road, for every new episode the car is spawned at random location with a random orientation. 
% We make different choices
% \noindent\textbf{Perception module.} 
% All the frames are stored in the buffer which is used to train the perception module while simultaneously training the policy.
% To train the VAE, 
Policy is trained as mentioned with a binary sparse reward if goal is completed before episode ends. 
The VAE module is trained in a standard self-supervised manner, with a binary cross entropy (BCE) reconstruction loss and a KL loss on the predicted distribution to enforce a normal in the encodings. 

While training perception and policy \textit{sequentially} is straightforward, we also investigate \textit{simultaneously} training both. 
% we investigate two strategies training either perception and policy \textit{simultaneously} or \textit{sequentially}.
In the latter case, all frames observed during policy training are only stored in a buffer from which data is sampled to optimize the VAE.\\
% When training simultaneously perception and control, 
% . For the encodings to follow a normal distribution, the kl divergence of the distribution predicted is also optimised. 
% This training is performed using data sampled from the buffer storing frames of each episodes for every different policies being trained simultaneously.\\

\begin{figure}
    \centering
    \includegraphics[width=1.0\linewidth]{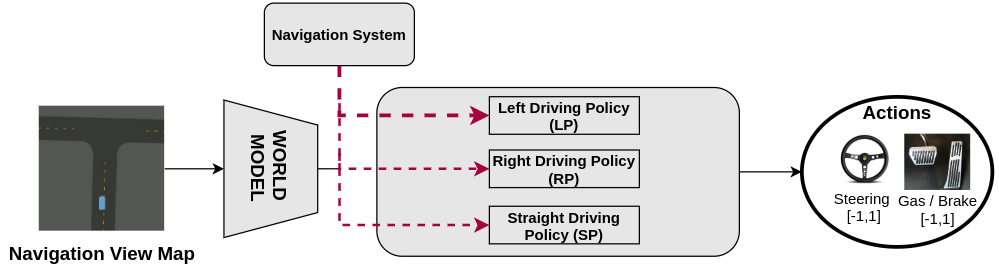}
    \caption{\textbf{Multi-policy strategy.} The navigation system switches policies according to the next intersection action.}
    \label{fig:Multi-policy}
\end{figure}

\noindent\textbf{Multiple policies.} 
Because driving is complex to learn with a single policy, we instead consider driving as tasks ensemble and learn multiple unique policies, switched at inference. In details, we learn three independent polices for driving \textit{straight} (SP), turning \textit{right} (RP), turning \textit{left} (LP). 
This not only helps simplifying the control module but also prevents integrating route information \textit{in} the world model.
% and also provide an external agent the ability to take high level decisions at the intersection. 
At inference we use an external navigation system (e.g. global planning algorithm) to switch between these policies near intersections, as shown in fig.~\ref{fig:Multi-policy}. 
% The episode starts by loading all the pre-trained weights for each of the sub policies. The agent starts with SP and continues as long as there are no intersection in the given frame. At the intersection an external agent (human) decides whether to switch to a new policy or continue with SP depending on its requirements. The navigation map provides a birds eye view hence the agent has enough time to plan. Here either human or a high level planning algorithm can be used as a navigation system.

% \begin{algorithm}
% \caption{Simultaneous learning of Perception and Control}
% \label{ppo-curr-algo}
% \begin{algorithmic}
% \State $\mathit{\pi_{\theta}, Q_{\theta}} \gets Kaiming\, Initialization$
% \State $\mathit{\pi_{\theta_{old}}, Q_{\theta_{old}}} \gets \pi_{\theta}, Q_{\theta}$
% \State $\mathit{E_{\theta}, D_{\theta} \gets Random\, Initialization}$
% \State $env \gets Initialise$
% \State $Goal \gets 0$
% \While {$Goal \neq 100$}
% \State Train VAE
% \State $s \gets env.reset(goal)$
% \For {$steps\, 1\, to\, 300$}
% \State $\mathit{action\, \thicksim {\pi_{\theta_{old}(s)}}}$
% \State $\mathit{s, r, done \gets env.step(action)}$
% \State $ppo\, buffer \gets store(s^{\prime}, action, s, r)$
% \State $vae\, buffer \gets store(s^{\prime})$
% \State $s \gets s^{\prime}$
% \If{done}
% \State break
% \EndIf
% \EndFor
% \State Train $\mathit{\pi_{\theta}}$
% \State $\mathit{\pi_{\theta_{old}}, Q_{\theta_{old}}} \gets \pi_{\theta}, Q_{\theta}$
% \State Clear ppo buffer
% \If{reward}
% \State $Goal \gets Goal + 1$
% \Else
% \State $Goal \gets Goal - 1$
% \EndIf
% \State Load latest VAE weights
% \EndWhile
% \end{algorithmic}

% \end{algorithm}

% \label{sec:method-test2}

\section{Experiments} \label{sec:experiments}
\label{sec:exp_metrics}

\begin{figure}
    \centering
    \footnotesize
	\setlength{\tabcolsep}{0.004\linewidth}
    \begin{tabular}{ccc}
        \includegraphics[height=0.27\linewidth]{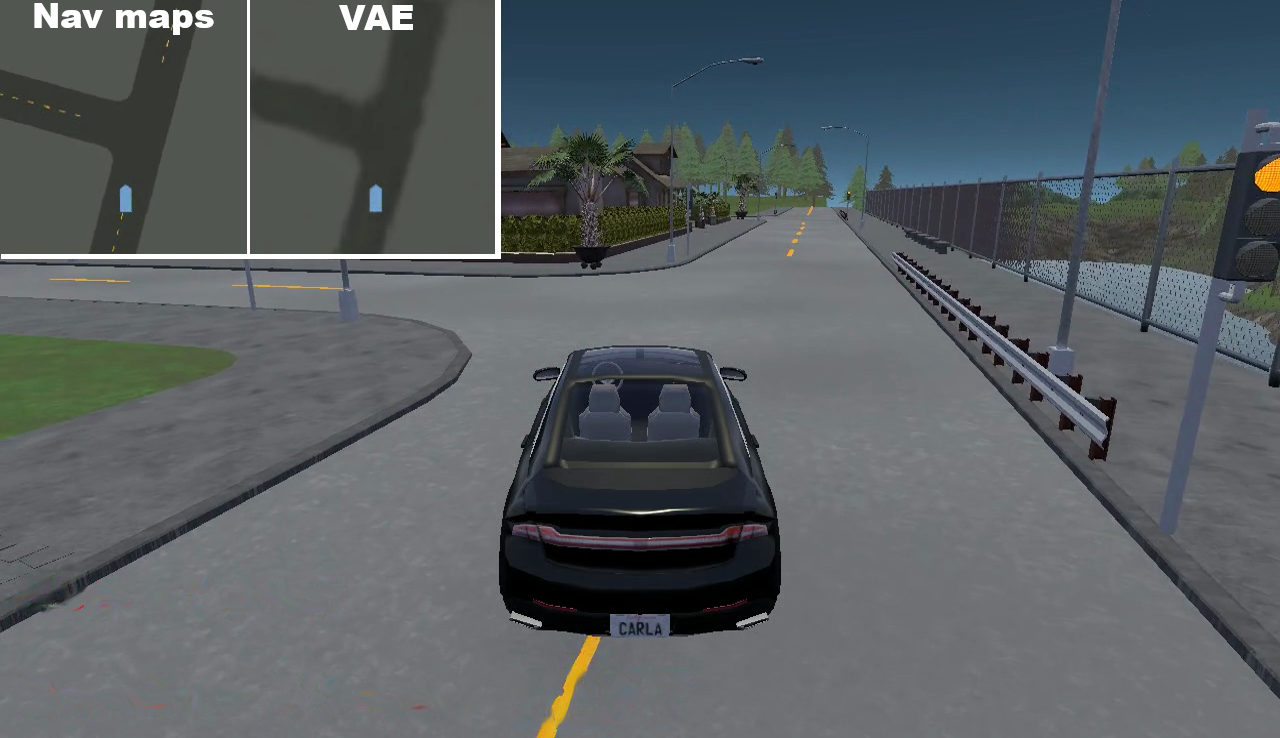} & \includegraphics[height=0.24\linewidth]{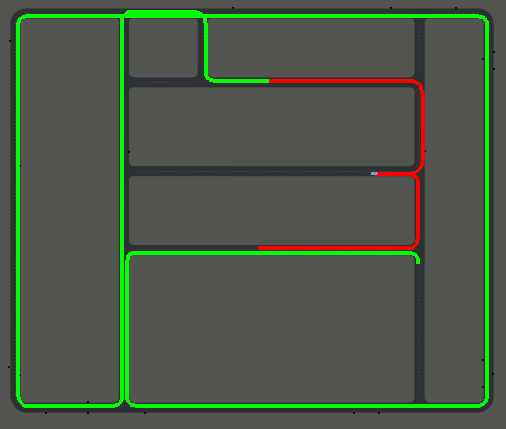} & \includegraphics[height=0.24\linewidth]{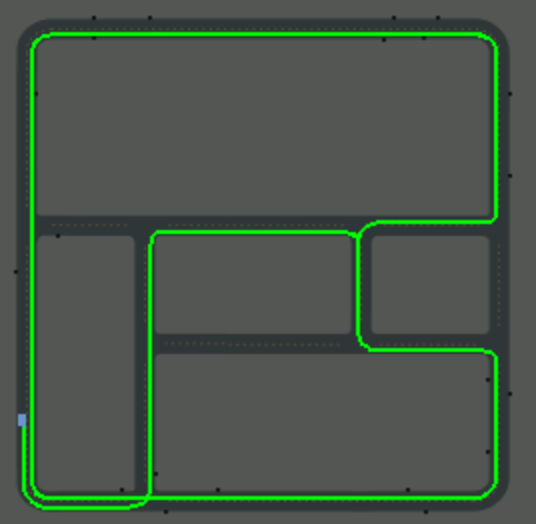} \\
        (a) 3D Carla simulator~\cite{Dosovitskiy2017CARLAAO} \ & \multicolumn{2}{c}{(b) Tracks layouts (Town 01, 02)}
    \end{tabular}
    \caption{\textbf{Simulator view and tracks layout.} (a) 3D simulator sample showing ego car with input maps and VAE as insets. Since we train with \textit{maps only}, we train on (b) a small track portion (red lines) and test on unseen areas (green lines).}
    \label{fig:tracks}
  \vspace{-1em}
\end{figure}

%\RC{Emphasize task difficult, mention no car.}
We conduct our experiments at~10Hz on Carla~\cite{Dosovitskiy2017CARLAAO} simulator, see Fig.~\ref{fig:tracks}a, as it provides a range of high definition maps for virtual towns, and custom API calls to interact with the simulator. 
Navigation views maps are custom-built and rendered as 256x256 RGB images.
Because these views have small domain gaps, we train only on three small sections of the track `Town01' which encompass the scenarios of interest (straight and left/right turns). Evaluation is conducted on test tracks `Town01' and `Town02', in unseen areas, see Fig.~\ref{fig:tracks}b. 
Considering the complexity of learning to drive from only maps input and sparse reward, our experiments are conducted in simulated environments having a single (ego) vehicle.
%\revcorr{}{Considering the complexity of learning a latent space and the policy with a sparse reward simultaneously, we limit our environment to a single ego vehicle only.}

In the following, we first provide details on the experimental setup, and evaluate the performance when perception and policy are trained simultaneously which takes as little as~8~hours on a single GPU. 
Finally, we ablate our contributions and compare simultaneous and sequential training. 
%\RC{Address why we don't compare to [Hafner et al., 2020] or D4PG,
%MPO, Soft Actor cretic}
% We report results when our pipeline simultaneously optimizes the perception and policy objectives. Simultaneously, it takes as little as 6-10hrs of training.

% Along with this the ease with which these maps can be improved according to our problem statement provided it an edge over other simulators. 
% Taking into account the brittle nature of training a reinforcement learning algorithm, the input map (Town 01) from the simulator are organised such that it is easy to differentiate between important decision making aspects like roads, side-walks and car. Along with this a user defined route is embedded onto the lanes. We first describe the experimental setup and then evaluate.

\subsection{Experimental Setup}
At each episode start, the agent are spawned with a random position on the road and orientation in $[-45\degree, 45\degree]$ w.r.t. the local road curvature.
 % We start our simultaneous training of multiple policies and perception module on three different routes comprising of left and right turns and straight. 
A buffer limit of 50000 is used and while the buffer is being filled, the perception and control modules optimizes the world model and policy, respectively. 
The VAE samples from the buffer and for every iteration optimizes the world model with a batch size of 100 and epochs equivalent to the number of samples batches. 
% The samples collected are 256 x 256 RGB images and are normalized to have values in the range of 0-1.

The episode duration $T$ (seconds) is a function of complexity, $T=\min(\max(c_i,10), 40)$. Short episodes for low complexity prevents exploration of unwanted states and early optimization of the policy. 
While clamping $T$ for larger complexity helps in optimizing the speed it also prevents unwanted wiggling, thus attaining the best policy. 
As optimizer we use Adam with a learning rate of 1e-5, and step-wise decay with step size of 5000 and $\gamma = 0.96$. The trajectories collected for each episode are trained for 15 epochs with a policy and value clip of 0.1. The discount factor used is 0.99.

\begin{figure}[]
  \centering
  
    \scriptsize
	\setlength{\tabcolsep}{0.01\linewidth}
    \begin{tabular}{ccccc}
        \textbf{Source} & \multicolumn{4}{c}{\textbf{Reconstruction}} \\
        \cmidrule(r){1-1}\cmidrule(lr){2-5}
        \includegraphics[height=0.18\linewidth,keepaspectratio]{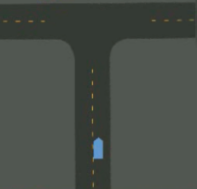} & 
        \includegraphics[height=0.18\linewidth,keepaspectratio]{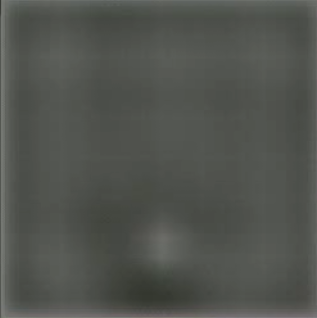} & \includegraphics[height=0.18\linewidth,keepaspectratio]{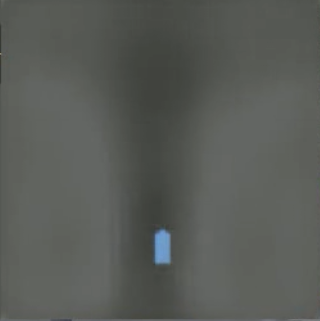} & \includegraphics[height=0.18\linewidth,keepaspectratio]{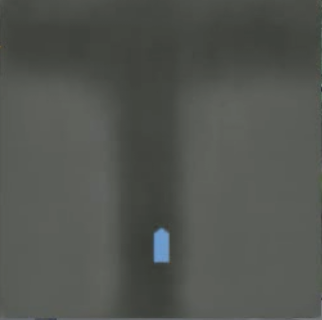} & \includegraphics[height=0.18\linewidth,keepaspectratio]{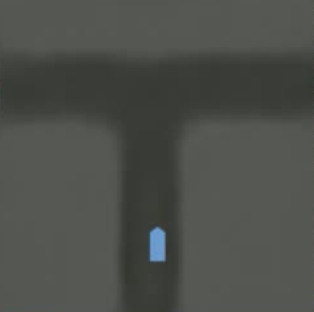}\\
        
        \includegraphics[height=0.18\linewidth,keepaspectratio]{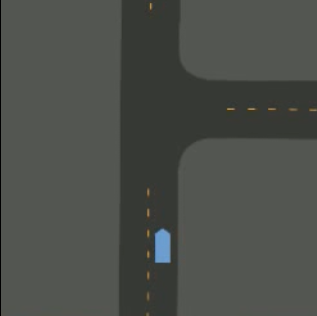} & 
        \includegraphics[height=0.18\linewidth,keepaspectratio]{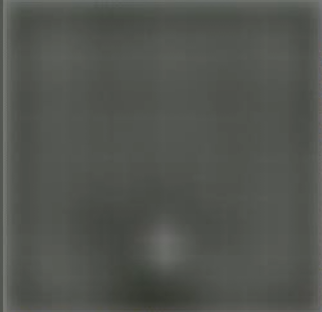} & \includegraphics[height=0.18\linewidth,keepaspectratio]{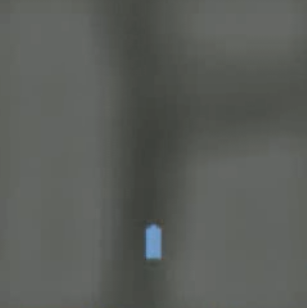} & \includegraphics[height=0.18\linewidth,keepaspectratio]{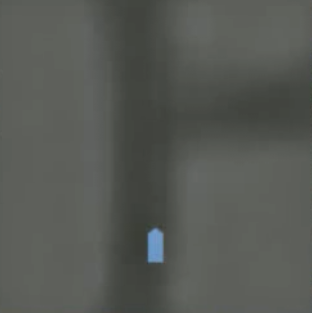} & \includegraphics[height=0.18\linewidth,keepaspectratio]{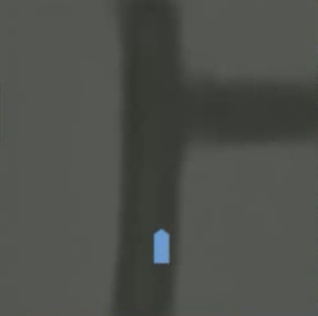}\\

        & 0 & 1000 & 2000 & 5000
    \end{tabular}
  \caption{\textbf{VAE reconstructions.} Qualitative reconstructions of source images along the training. The VAE quickly converges and discover main scene traits around 1000 training steps.}
  \label{fig:vae}
  \vspace{-1.5em}
\end{figure}

\begin{figure}
    \centering
    \subfloat[VAE loss]{\includegraphics[width=0.48\linewidth]{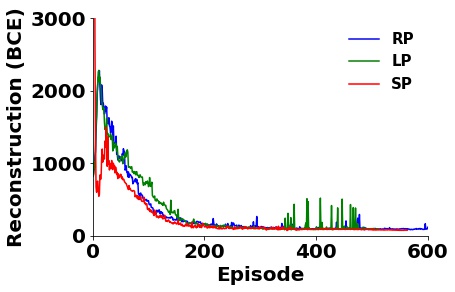}\label{fig:graphs_simult_vae}}\hfill
    \subfloat[Policies curriculum]{\includegraphics[width=0.48\linewidth]{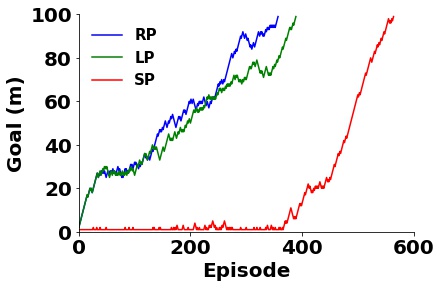}\label{fig:graphs_simult_complexity}}
    \caption{\textbf{Simultaneous Perception and Policy training.} (a)~VAE reconstruction loss quickly converges after only 200 episodes, while \textit{simultaneously} the three parallel policies~(b)~are trained on increasing curriculum complexity.\\}
    \label{fig:graphs}
    \vspace{-1.5em}
\end{figure}

\subsection{Performance}
We evaluate our method proposal in threefold, first briefly evaluating the perception module to ensure effectiveness of our world model, second evaluating the policy performance on train tracks, and third studying the generalization capacity of our world model and policies.\\

\noindent\textbf{Perception.} We qualitatively evaluate the world model learned by the VAE, showing reconstructions in Fig.~\ref{fig:vae}. From the latter, we note that the reconstruction improves during the training, and that only a few thousand iterations is sufficient to restore the global road layout. Quantitatively, the reconstruction loss during training in Fig.~\ref{fig:graphs_simult_vae} also demonstrates the quick BCE loss convergence.\\

\noindent\textbf{Policy.} To evaluate policy, Fig.~\ref{fig:graphs_simult_complexity} shows the goal distances in meters --~i.e. curriculum complexity~-- for all three policies trained in parallel. The turn left policy (LP) and turn right policy (RP) evolve concurrently with visible effect of our curriculum \textit{revert} strategy since the goal partly decrease locally. 
Conversely, we denote that the straight policy (SP) requires significantly more episodes to optimize. We conjecture that it relates to the overcompensation problem of RL agents~\cite{Jaritz2018EndtoEndRD}.
Ultimately, all policies converged in less than 600 episodes, with agents driving at an average speed of 10.5kmph. 
Since sparse reward does not enforce speed, the velocity in fact relates to the minimum acceptable speed to complete episodes being 9kmph (i.e. the episode duration $T$ forces the agent to complete 100m in less than 40sec). We conjecture that decreasing the episode duration could further speed up the driving.

\begin{figure}
  \centering
    \scriptsize
	\renewcommand{\arraystretch}{0.6}
	\setlength{\tabcolsep}{0.003\linewidth}
    \begin{tabular}{cccc}
        & \textbf{Left Policy (LP)} & \textbf{Straight Policy (SP)} & \textbf{Right Policy (RP)}\\
        \rotatebox{90}{~~~~~~~~~\textbf{Train}}\hspace{2pt}  & \includegraphics[width=0.3\columnwidth,height=0.30\columnwidth,]{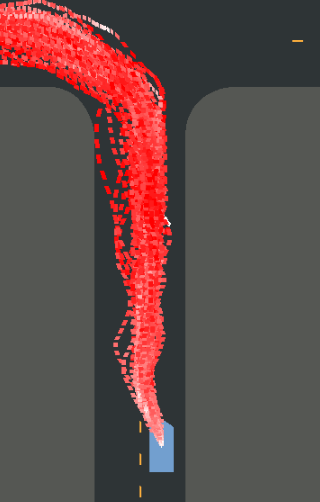} & \includegraphics[width=0.3\columnwidth,height=0.30\columnwidth,]{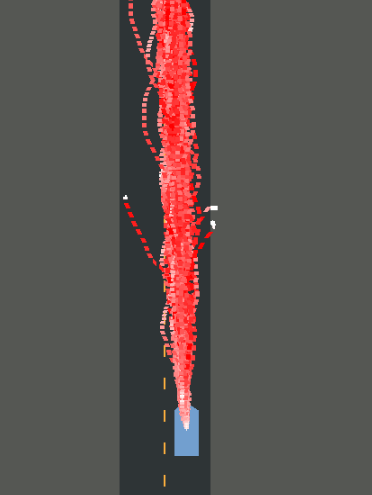} & \includegraphics[width=0.3\columnwidth,height=0.30\columnwidth,]{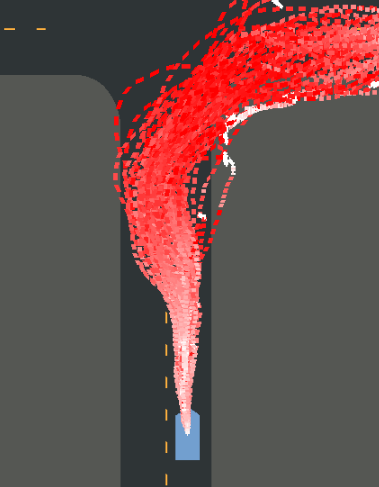} \\
        \rotatebox{90}{~~~~~~~~~\textbf{Test}} & \includegraphics[width=0.3\columnwidth,height=0.30\columnwidth,]{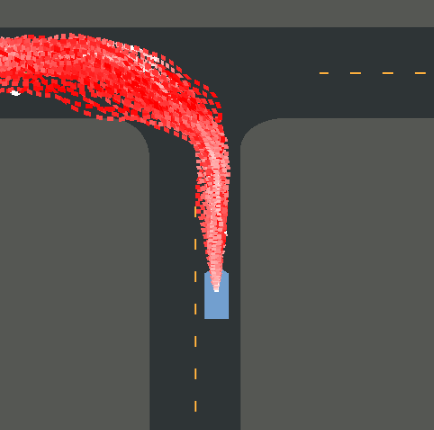} & \includegraphics[width=0.3\columnwidth,height=0.30\columnwidth,]{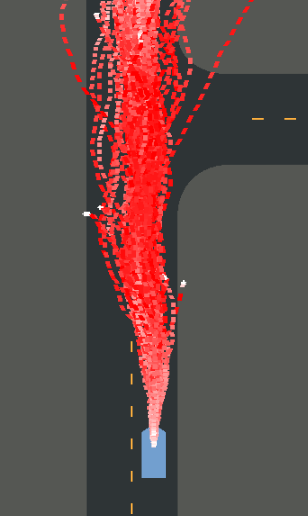} & \includegraphics[width=0.3\columnwidth,height=0.30\columnwidth,]{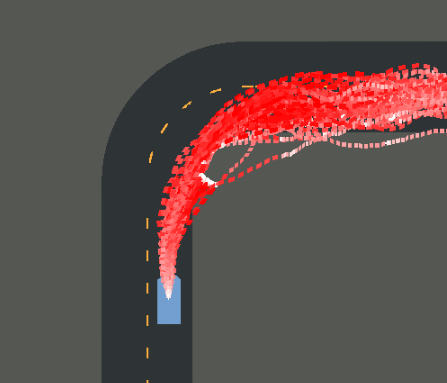} \\
        % Ours (test) & & & &  \\
    \end{tabular}

 \caption{\textbf{Driving style variability.} Visualization of 100 runs for a given goal, for train tracks (Town 01) (top) or unseen test tracks (Town 01 and 02) (bottom). Our method performs similarly or either setup related to the little domain gap, and discovers proper driving behavior despite weak binary supervision. Normalized speed (white$\mapsto$red) shows agents learned to slow down before turns.}
 \label{fig:labe20}
% \vspace{-1em}
\end{figure}

On driving style, Fig.~\ref{fig:labe20} \textit{top} shows 100 runs when driving either policy (left, straight, right) on training tracks, with markers color evolving from high speed (red) to low speed~(white). On train tracks, despite the absence of any dense reward, we denote the agents successfully learned to drive in the safe drivable area. Even more, the car naturally stays close to the track center although wiggling is visible as often mentioned in the literature~\cite{wolf2017learning,Jaritz2018EndtoEndRD}. 
Finally, on \textit{Left/Right Policy} we observe that the agents learned to turn from the inside by shifting on the right or left first. While this lead to driving rules infringement we highlight that such time-optimized driving is impossible with dense rewards that penalize distance from the lane center~\cite{Jaritz2018EndtoEndRD}.\\

\noindent\textbf{Generalization.} Qualitative driving on unseen test tracks, Fig.~\ref{fig:labe20} \textit{bottom}, also demonstrates generalization of the learned policies and the world model. 
Despite the small domain gap of our navigation views, notice that test tracks include \textit{unseen} road layout which both the world model and policy have to cope with. This is visible on test tracks for example in straight policy (SP) and right policy (RP), Fig.~\ref{fig:labe20}~\textit{bottom}.
% images shows unseen road layout which the agent is capable to cope with.

To further study generalization, Tab.~\ref{tab:succesrate} reports the success rate for different goal distances on \textit{test tracks only} (for 100 runs). 
From the latter, \textit{ours simultaneous} completes 20m goal with 100\% success rate, and 100m goal -- the highest trained complexity -- with 82\% success rate. Of interest, we demonstrate that despite that agents only train with up to 100m goals, the policy scales to long driving goals like 200m~(66\%) and 300m (41\%). 
Training perception and policy pipeline in a sequential manner (\textit{ours sequential}), we denote few percent better success rates. However, this comes at the cost of more complex and twice longer training ($\sim$18~hours), which is why we prefer simultaneous training. In the supplementary video we show our agent -- although only trained up to 100m -- is able to generalize to unseen tracks and maneuvers much longer distance ($\sim$1.2kms).
%  \RC{Highlight better test bigger than 100m}

% show the visualisation of the final policy on both train and test route. Considering this and performance on unseen long routes ~\ref{tab:succesrate}, the final policy generalises to unseen tracks.

\captionsetup[subfigure]{labelformat=empty}

\begin{table}[]
    \scriptsize
    \centering
	\setlength{\tabcolsep}{0.033\linewidth}
    \begin{tabular}{cccc|cc}
    \toprule
        & \multicolumn{5}{c}{Success rate} \\
        \cmidrule(lr){2-6}
        Method & 20m & 50m & 100m & 200m & 300m \\% & Avg. speed (km/h) \\
        \midrule
        \textbf{Ours - simultaneous} &\textbf{1.0} &0.85 &0.82 &0.66 &0.41 \\%&10.5   \\
        \textbf{Ours - sequential} & 0.91 &\textbf{0.90} &0.90 &\textbf{0.69} &\textbf{0.51} \\%12.5 \\
        \midrule
        w/o revert &0.51 &0.08 &0.02 &0 &0 \\%4.7\\
        w/ curr. +2 &0.99 &0.89 &\textbf{0.91} &0.54 &0.07 \\%\textbf{13.8}\\
        w/ fixed episode dur. &0.36 &0.04 &0 &0 &0 \\%6.7\\
        w/o constraints &0.87 &0.44 &0.11 &0 &0 \\%12.44 \\
        \bottomrule
    \end{tabular}
    \caption{\textbf{Driving test tracks performance.} Success rate of goal completion on \textit{unseen test tracks} for different goal distances. The first two lines denote our pipeline with perception and policy either trained simultaneous or sequential, while bottom lines are ablations of our contributions. Even if training does not exceed 100m, policy is able to drive longer.}
    \label{tab:succesrate}
    \vspace{-1.5em}
\end{table}

\subsection{Ablation}
\begin{figure}
    \centering
    \subfloat[(a) Individual/All policies]{\includegraphics[width=0.38\linewidth,height=0.23\linewidth]{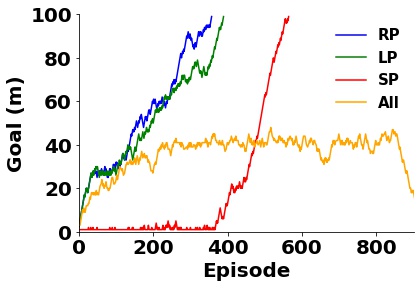}\label{fig:graphs_policies_complexity}}
    \subfloat[(b) Goal constraints]{\includegraphics[width=0.32\linewidth]{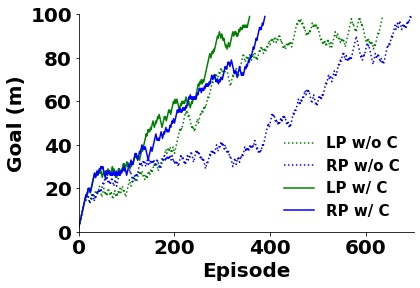}\label{fig:graphs_constraints_complexity}\includegraphics[width=0.32\linewidth]{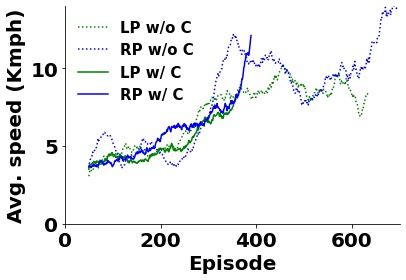}\label{fig:graphs_constraints_speed}}
    \caption{(a) Effect of learning individual policies (\textit{LP,SP,RP}) or as a single policy (\textit{All}). (b) Comparing the performance of different policies with and without goal constraints.}
    \label{fig:graphs_ablation}
%    \vspace{-1.5em}
\end{figure}

We now discuss ablation of our contributions, and report quantitative performance in Tab.~\ref{tab:succesrate} and Fig.~\ref{fig:graphs_ablation}.\\ 

\noindent\textbf{Individual policies vs Multiple policies.}
To study the effect of learning multiple task simultaneously, using sparse rewards, we compare in Fig.~\ref{fig:graphs_policies_complexity} the performance of our pipeline when training either indivual policies (left, straight, right) separately or all together in a single policy. From the plot, it is clearly inefficient to learn all policies (`All') in comparison to individual policies (`LP', `SP', `RP'). 
Specifically, when learning all policies grouped, it optimizes slowly and reaches a plateau (around 40m complexity), while our individual policies all reaches max goal complexity (100m).

% \paragraph{Sequential vs Simultaneous training} While simultaneous training of perception and control module is challenging considering the non stationary distribution of latent space it shows similar performance to sequential training as evident from Table ~\ref{tab:succesrate}. \RC{2x faster}

\noindent\textbf{Goal constraints.} 
We evaluate the benefit of our goal constraints described in Sec.~\ref{sec:meth-sparsecur}.
From Fig.~\ref{fig:graphs_constraints_complexity}, it is clear that without constraints (`w/o C', dotted) the policy still manages to achieve similar curriculum complexity, though speed graph shows that without constraints the agent tend to drive slower. Also, from Tab.~\ref{tab:succesrate}, without constraints the success rate is in fact significantly slower, and is even unable to generalize for long distance goals (over 100m).
% External supervisions are added, for better driving. These includes distance from the centre and orientation of the car. 
In the sparse reward settings, these constraints are indeed useful when an agent needs to perform a sub task along with the main goal (like driving following the traffic rules). 
% It is interesting to observe the policies without any guidance tends to achieve optimal performance earlier by driving safe at a lower speed. 
The addition of constraints prevents the agent from exploring different actions leading to sub-optimal performance. The agent without constraints tends to deviate a lot initially in the training phase even for lesser complexity goal but learns the optimal actions with time.

\noindent\textbf{Curriculum strategies.} We also evaluate the benefit of our curriculum strategy in Tab.~\ref{tab:succesrate}. Performance without our \textit{revert strategy} (`w/o revert') demonstrate the significant benefit of our yet simple strategy, as it avoids remaining stuck in a local minimum.
Increasing complexity twice faster (`w/ curr +2') show acceptable but worse performance.
% Our proposed curriculum strategy can be tuned to have a different rate of increment/decrement, it is observed from Table ~\ref{tab:succesrate} different approach leads to poor optimization to the best policy. 

%\subsection{Performance}
% \captionsetup[subfigure]{labelformat=empty}
%!TEX root = main.tex

\section{Conclusion}
\label{sec:concl}

We propose the first sparse reward dependent reinforcement learning agent for end to end driving. Instead of front view images, we rely on navigation maps to learn individual policies in parallel and simultaneously train perception and world model.
% To counter the challenges of sparse signal we use a distance based curriculum strategy for efficient training. This is challenging and to take into account the different tracks we introduce a multi policy strategy learning independent policies for left turn, right turn and straight driving. This approach transfers the control to an external agent for high level decisions at the intersection. The performance of the final policy on unseen tracks show that driving behaviour can be learnt without external supervision in the form of dense reward.
Our results demonstrate exciting generalization capacity and natural emergence of driving styles. We also believe RL driving with constrained sparse rewards open doors to new perspective such as the inclusion of driving rules -- a major RL challenge especially when accounting for other road users.
%\revcorr{}{and several other complex tasks like driving without collision or overtaking in the presence of multiple vehicles as different constraints}. 
% \RC{mention adding objects}
% We think that this goal conditioned sparse reward strategy can enable learning different driving skills like approaching a traffic light, yielding, overtaking using simple sparse reward setting.

{\small
\bibliographystyle{ieee}
\bibliography{biblio}
}

\end{document}